\documentclass[10pt,twocolumn,letterpaper]{article}

\usepackage{wacv}              % To produce the CAMERA-READY version
\usepackage{graphicx}
\usepackage{amsmath}
\usepackage{amssymb}
\usepackage{booktabs}
\usepackage{algorithmicx,algorithm,algpseudocode}
\usepackage{adjustbox}
\usepackage{xcolor}

\usepackage{bbm}
\usepackage[flushleft]{threeparttable}

\newcommand{\beginsupplement}{%
    \clearpage
    \onecolumn{
        \centering
        \Large
        \textbf{Multi-view Image Diffusion via Coordinate Noise and Fourier Attention}\\
        \textbf{(Supplementary Material)}\\
        \vspace{1em}
        \large
        \text{Justin Theiss
        \quad
        Norman M\"uller
        \quad 
        Daeil Kim 
        \quad 
        Aayush Prakash}\\
        \text{Meta Reality Labs}\\
    }
    \setcounter{section}{0}
    \renewcommand{\thesection}{\Alph{section}}
    \setcounter{table}{0}
    \renewcommand{\thetable}{S\arabic{table}}%
    \setcounter{figure}{0}
    \renewcommand{\thefigure}{S\arabic{figure}}%
    \setcounter{equation}{0}
    \renewcommand{\theequation}{S\arabic{equation}}%
}

\usepackage[pagebackref,breaklinks,colorlinks]{hyperref}

\usepackage[capitalize]{cleveref}
\crefname{section}{Sec.}{Secs.}
\Crefname{section}{Section}{Sections}
\Crefname{table}{Table}{Tables}
\crefname{table}{Tab.}{Tabs.}

\def\wacvPaperID{1484} % *** Enter the WACV Paper ID here
\def\confName{WACV}
\def\confYear{2025}

\begin{document}

\title{Multi-view Image Diffusion via Coordinate Noise and Fourier Attention}

\author{Justin Theiss
\qquad
Norman M\"uller
\qquad 
Daeil Kim 
\qquad 
Aayush Prakash\\
Meta Reality Labs\\
}
\maketitle

\begin{abstract}
Recently, text-to-image generation with diffusion models has made significant advancements in both higher fidelity and generalization capabilities compared to previous baselines. However, generating holistic multi-view consistent images from prompts still remains an important and challenging task. To address this challenge, we propose a diffusion process that attends to time-dependent spatial frequencies of features with a novel attention mechanism as well as novel noise initialization technique and cross-attention loss. This Fourier-based attention block focuses on features from non-overlapping regions of the generated scene in order to better align the global appearance. Our noise initialization technique incorporates shared noise and low spatial frequency information derived from pixel coordinates and depth maps to induce noise correlations across views. The cross-attention loss further aligns features sharing the same prompt across the scene. Our technique improves SOTA on several quantitative metrics with qualitatively better results when compared to other state-of-the-art approaches for multi-view consistency.
\end{abstract}

\section{Introduction}
\label{sec:intro}

\begin{figure*}[ht]
\centering
\includegraphics[width=0.98\linewidth]{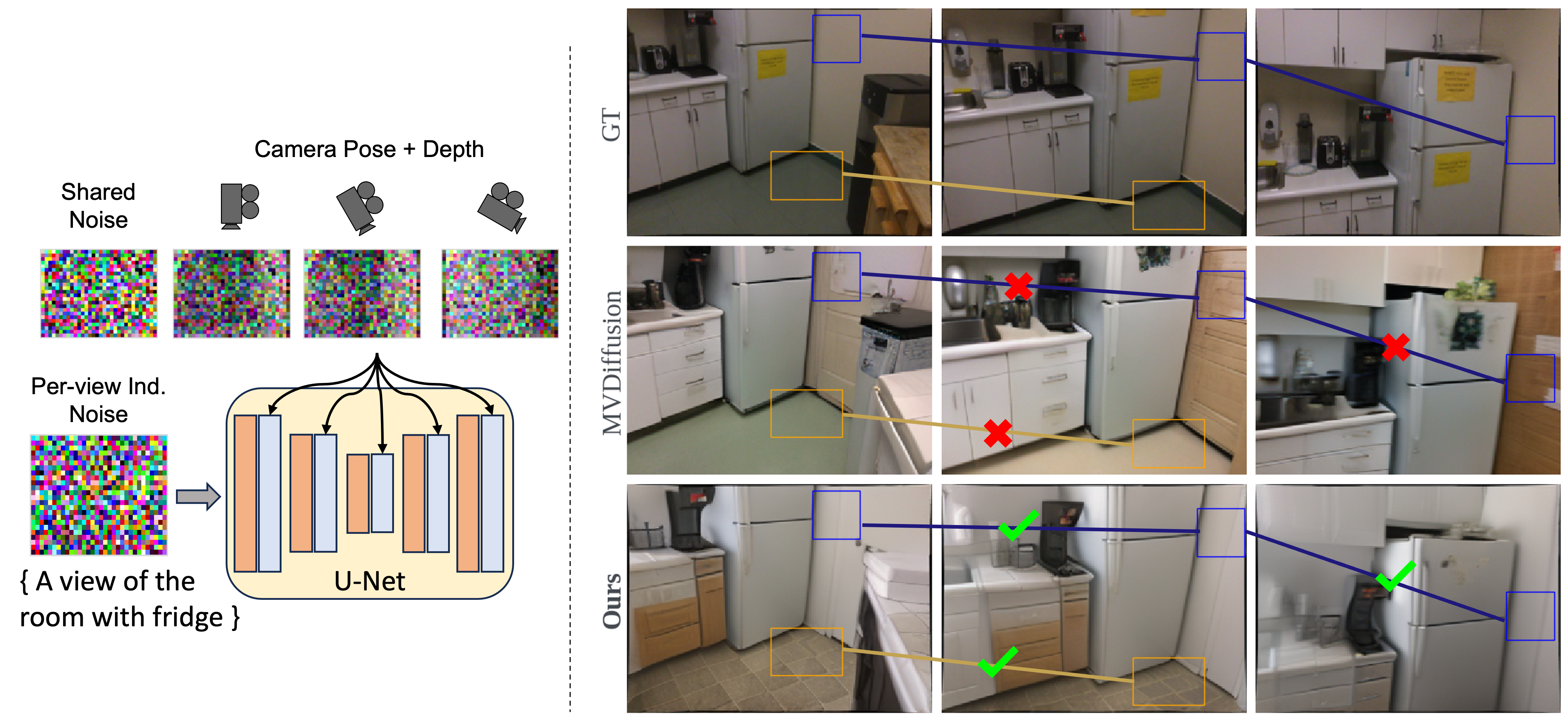}
\caption{We propose a method that addresses the lack of consistency in multi-view image generation by aligning appearance in non-overlapping regions of multi-view scenes (\textit{left}). Compared to MVDiffusion, our approach improves the consistency of textures and geometry, particularly in non-overlapping regions (\eg, floor and walls; \textit{right}).}\label{fig:teaser}
\end{figure*}

In recent years, significant breakthroughs have been made in text-conditional image generation~\cite{rombach2022LDM,ramesh2022hierarchical,saharia2022photorealistic,zhang2023adding, nichol2021glide,rombach2022LDM}. However, when extending single-view image generation to multi-view and video generation from text prompts \cite{tang2023mvdiffusion, lee2023syncdiffusion, bar2023multidiffusion, hollein2023text2room, ge2023preserve, geyer2023tokenflow, wu2023freeinit, gu2023reuse, qiu2023freenoise, ren2024consisti2v} there remains considerable challenges, particularly around the consistency of a scene's geometry and appearance. 
To this end, recent works~\cite{bar2023multidiffusion,tang2023mvdiffusion, blattmann2023align} implement attention modules that process all views simultaneously \cite{tang2023mvdiffusion, blattmann2023align}. This aims to align features across views by incorporating cross-attention modules into the standard diffusion model architecture. 
Moreover, MVDiffusion \cite{tang2023mvdiffusion} uses known camera pose and depth information in order to find corresponding points for attention across different views. Similarly, ConsistI2V~\cite{ren2024consisti2v} proposed changes to cross attention between views, such as attending to the local neighborhood around a query index for each view, but performance seems constrained to video sequences with high temporal sampling. In the case of more general multi-view image generation (\eg, panoramas), such a method may not adequately handle larger changes in camera pose between views. Specifically, for methods relying on high overlap between frames, the appearance in areas with less overlap across the scene often exhibit stark changes (see Figure \ref{fig:teaser}). Improving consistency in non-overlapping regions is therefore important for ensuring consistency in the global appearance.

Another exciting direction to improve the multi-view consistency of text-to-image is through the role of noise initialization by using shared noise \cite{ge2023preserve}, correlated noise \cite{qiu2023freenoise}, or low spatial frequency components of images \cite{wu2023freeinit}. These studies have shown that overall appearance can be improved by combining shared and independent components when initializing noise for multi-view generation. One possible explanation for this effect is the recently observed gap between noise used during training and inference \cite{lin2024common}. At the noisiest time step during training, low spatial frequency information regarding the image is still present; however, during inference, this information is missing when sampling from Gaussian noise. In this work, we leverage this initialization gap in order to improve consistency across generated images by inducing low-frequency correlations across noise samples without requiring access to ground truth images \cite{ren2024consisti2v} or costly sampling steps to generate a starting layout \cite{wu2023freeinit}.

We address the challenge of multi-view consistent text-to-image generation through a novel diffusion-based method that combines noise initialization with Fourier-based attention to guide image generation toward a consistent appearance. 
Building on recent work highlighting the gap in signal-to-noise ratio between noise samples used during training and inference, we propose a method for coordinate-based noise initialization that induces low spatial frequency correlations in the noise samples across views. We further propose an attention module that aligns non-overlapping regions across views by attending to progressively higher spatial frequency features across denoising time steps. Finally, we introduce a prompt-based cross-attention loss that ensures attention between prompt tokens and each view is consistent with the ground truth scene. Overall, our method improves multi-view consistency by initializing with pose-dependent noise, attending to frequency-dependent features in non-overlapping regions, and ensuring consistent semantic relationships via a prompt-based cross-attention loss.

In order to understand the effect of various design choices for more disparate views, we evaluate performance for multi-view consistency using two settings: panoramic and depth-conditioned image generation. We demonstrate quantitatively and qualitatively that these design choices improve image quality and multi-view consistency over state-of-the-art approaches. We summarize our contributions as follows:

\begin{itemize}
    \item We introduce a novel noise initialization technique, which incorporates shared noise and low spatial frequency information across views without time-prohibitive diffusion inversion or access to real images.
    \item We introduce a novel attention mechanism (Fourier-based Attention) that attends to shared-noise features across non-overlapping regions of a scene.
    \item Finally, we introduce a novel cross-attention loss that aligns multi-view prompt cross-attention maps with the ground truth attention maps, improving alignment of features sharing the same prompt across views.
\end{itemize}

\section{Related Work}
\label{sec:related_work}

\subsection{Text-to-Image Diffusion Models}

The field of text-to-image diffusion models has undergone considerable advancements, with significant contributions from models like DALL-E 2 \cite{ramesh2022hierarchical}, GLIDE \cite{nichol2022glide}, Latent Diffusion Models (LDMs) \cite{rombach2022LDM}, and Imagen  \cite{saharia2022photorealistic}. These models excel in generating photo-realistic images from text prompts, combining the efficiency of large-scale diffusion models with the sophistication of pre-trained language models. Additional control over the image output is possible through manipulation of model cross-attention layers, as demonstrated in Prompt-to-Prompt \cite{hertz2022prompt}, Attend-and-Excite \cite{chefer2023attend}, and FreestyleNet \cite{xue2023freestyle}. This cross-attention control has also lead to improvements in multi-view consistency. Our model extends the single-view text-to-image LDM into the multi-view domain with additional conditioning on camera pose and/or depth.

\subsection{Multi-view Consistency in Image Diffusion Models}
The pursuit of multi-view consistency in image generation has led to several noteworthy advancements. MultiDiffusion \cite{bar2023multidiffusion} focused on fusing diffusion paths for controlled image generation, addressing the seamless integration of multiple views. SyncDiffusion \cite{lee2023syncdiffusion} synchronizes joint diffusions for coherent montage creation, using gradient descent from perceptual similarity loss to align multiple diffusions. DiffCollage \cite{zhang2023diffcollage} generates large content by merging results from overlapping nodes represented by a factor graph.  However, all these techniques either do not generate true panorama (left and right corners mismatch) or have visible artifacts in the generated images. 
TokenFlow\cite{geyer2023tokenflow} generates multi-view consistent edits in videos by propagating features based on inter-frame correspondence, but needs access to ground truth noise samples.
MVDiffusion \cite{tang2023mvdiffusion} improved multi-view image generation by embedding correspondence-aware attention in diffusion models, optimizing for consistency across multiple views. However, MVDiffusion struggles to generate multi-view consistent images in non-overlapping regions. Our method explicitly attends to such regions to align appearance.

\subsection{Noise Initialization in Diffusion Models}
Recent works have investigated the gap in noise initialization between training and inference for diffusion models \cite{lin2024common, zhang2024preserving}, noting that there is an information leak that occurs even at highest noise levels during training. Other works have leveraged this information leak as a way to improve consistency in appearance by incorporating low spatial frequency information \cite{wu2023freeinit, ren2024consisti2v} or the mean of images from a given class \cite{zhang2024preserving}. In video diffusion models, initializing with a weighted combination of shared and independent noise across frames \cite{ge2023preserve} or inducing long-range correlations via noise rescheduling \cite{qiu2023freenoise} have similarly achieved greater global consistency. Our method instead leverages 3D coordinate information to inform \emph{spatial} structure of the scene.
\section{Method}
The method section is organized as follows: we first cover the preliminaries regarding diffusion models, then we propose our method for noise initialization, Fourier-based attention, and prompt-based cross-attention loss. We end the section with a description of the full training paradigm.

\label{sec:method}

\begin{figure*}[ht]
\centering
\includegraphics[width=0.98\linewidth]{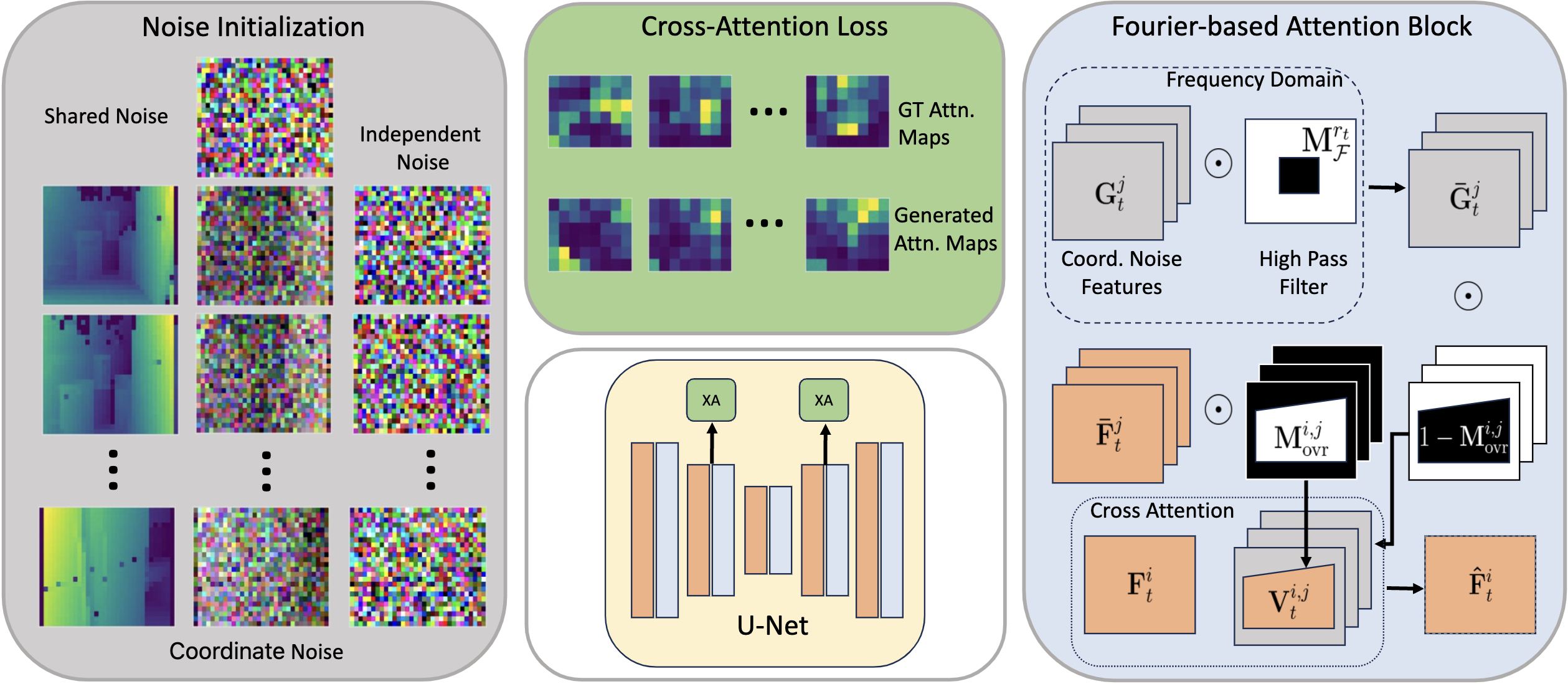}
\caption{Overview of our proposed technique. (\textit{Left}) We initialize noise by sampling Gaussian noise shared across views as well as independent per view. We combined the shared noise with depth or transformed pixel coordinates to obtain \emph{Coordinate Noise} (Sec. \ref{sec:noise_init}), which provides a low spatial frequency bias to inform the overall structure of the scene. (\textit{Center}) We add our Fourier-based Attention (FBA) blocks (blue, see \textit{right} panel) within the U-Net architecture and introduce a cross-attention loss (Sec. \ref{sec:xa_loss}) to ensure consistent spatial relationships across views. (\textit{Right}) Finally, our novel attention module (Sec. \ref{sec:fba}) time-dependent spatial frequencies of features generated from the \emph{Coordinate Noise} in non-overlapping regions to better align the global appearance across the scene.}\label{fig:method}
\end{figure*}

\subsection{Preliminary}
\label{sec:prelim}
Image diffusion models are trained to model a data distribution $p_{data}$ by iteratively denoising an image $\mathbf{x}$ from a random Gaussian noise sample across a sequence of $T$ time steps. Latent Diffusion Models (LDMs) instead operate on a latent representation $\mathbf{z}$ from a pre-trained VAE autoencoder (\ie, $\mathbf{z} := \mathcal{E}(\mathbf{x})$ and $\mathbf{x} := \mathcal{D}(\mathbf{z})$). During the forward diffusion process, noise is added to the latent $\mathbf{z}_0$ at each time step $t$:

\begin{equation}
    q(\mathbf{z}_t | \mathbf{z}_{t - 1}) = \mathcal{N}(\mathbf{z}_t; \sqrt{\alpha_t} \mathbf{z}_{t - 1}, \sigma^{2}_{t} \mathbf{I}),
    \label{eqn:forward_diffusion}
\end{equation}

\noindent where the noise schedule is defined by parameters $\alpha_t$ and $\sigma^{2}_{t}$ derived from a predefined variance schedule $\beta_{1}, \ldots, \beta_{T}$ with $\sigma^{2}_{t} = \beta_t$ and $\alpha_t = 1 - \beta_t$. In practice, the forward diffusion can be determined in a single step:

\begin{equation}
    q(\mathbf{z}_t | \mathbf{z}_0) = \mathcal{N}(\mathbf{z}_t; \sqrt{\bar{\alpha}_t} \mathbf{z}_0, (1 - \bar{\alpha}_t) \mathbf{I})
    \label{eqn:forward_diffusion_direct}
\end{equation}

\noindent where $\bar{\alpha}_t = \prod^{t}_{i}{\alpha_{i}}$. Therefore, the noisy latent $\mathbf{z}_t$ can be directly sampled with Gaussian noise $\epsilon \sim \mathcal{N}(0, \mathbf{I})$:

\begin{equation}
    \mathbf{z}_t = \sqrt{\bar{\alpha}_t} \mathbf{z}_0 + \sqrt{1 - \bar{\alpha}_t} \epsilon.
    \label{eqn:forward_noise_step}
\end{equation}

The LDM is then trained to approximate the reverse diffusion process, in order to obtain the latent $\mathbf{z}_0$ from Gaussian noise $\mathbf{z}_T$:

\begin{align}
    p_{\theta}(\mathbf{z}_{0:T}) &= p(\mathbf{z}_T) \prod^{T}_{t = 1}{p_{\theta}(\mathbf{z}_{t - 1} | \mathbf{z}_t)} \\
    p_{\theta}(\mathbf{z}_{t - 1} | \mathbf{z}_t) &= \mathcal{N}(\mathbf{z}_{t - 1}; \mu_{\theta}(\mathbf{z}_t, t), \Sigma_{\theta}(\mathbf{z}_t, t)),
    \label{eqn:reverse_diffusion}
\end{align}

\noindent where $\mu_{\theta}$ and $\Sigma_{\theta}$ are predicted by the denoising network $\epsilon_{\theta}$, typically implemented using the U-Net \cite{ronneberger2015u} architecture. The denoising network $\epsilon_{\theta}$ is then trained to predict the ground truth noise $\epsilon$ by minimizing the following objective function:

\begin{equation}
    \mathcal{L}_{LDM} = \mathbb{E}_{t, \mathbf{z}_0, \epsilon} {\lVert \epsilon - \epsilon_{\theta}(\mathbf{z}_t, t) \rVert}^{2}_{2}.
    \label{eqn:ldm_loss}
\end{equation}

\subsection{Coordinate-based Noise Initialization}
\label{sec:noise_init}
During inference, the noisy latent $\mathbf{z}_T$ is sampled from $\mathcal{N}(0, \mathbf{I})$ and the denoised sample $\mathbf{z}_0$ is decoded using the pre-trained decoder $\mathcal{D}$ to obtain the generated image $\hat{\mathbf{x}}$. However, as noted in recent works \cite{lin2024common, zhang2024preserving}, there is an SNR gap between the noisy latent $\mathbf{z}_T$ used during training and inference that leads to a information leak provided to the model during training but not during inference. This suggests that consistency in multi-view image generation may be improved by incorporating shared noise and/or low spatial frequencies from a target image when initializing the noise latent $\mathbf{z}_T$. Indeed, rather than using the latent sampled from $\mathcal{N}(0, \mathbf{I})$, previous works have observed that incorporating latent features from real images \cite{wu2023freeinit, ren2024consisti2v} into $\mathbf{z}_T$ can improve the consistency when generating multiple images (\eg, a video sequence). In our experiments, we explore the use of noise initialization methods that do not require access to the diffusion-inverted latent features $\mathbf{z}_0$ of real images during inference.

Based on the aforementioned gap between the latent $\mathbf{z}_T$ used during training and inference, it is intuitive to initialize noise using Equation \ref{eqn:forward_noise_step} by replacing the original latent $\mathbf{z}_0$ with some shared noise or image features $\mathbf{\epsilon}_{\text{shared}}$:

\begin{equation}
    \mathbf{\hat{z}}^{i}_T = \sqrt{\bar{\alpha}_T} \mathbf{\epsilon}_{\text{shared}} + \sqrt{1 - \bar{\alpha}_T} \mathbf{\epsilon}^{i},
    \label{eqn:shared_noise_init}
\end{equation}

\noindent where $\mathbf{\epsilon}_{\text{shared}}$ is shared across all views and $\mathbf{\epsilon}^{i} \sim \mathcal{N}(0, \mathbf{I})$ is sampled independently per view. Setting $\mathbf{\epsilon}_{\text{shared}}$ to random Gaussian noise has the effect of providing similar shared statistics across views, but does not necessarily provide information about the scene generally.

In order to provide information about the scene, we propose incorporating depth maps (where available) and transformed pixel coordinates based on camera pose information between each view and a reference (\eg, center view of a panorama). By utilizing the camera pose and (optionally) depth information of the scene, we are able to induce low spatial frequency correlations implicitly across the scene. We then combine the coordinate-based component with the shared noise to obtain our noise initialization:

\begin{align}
    \mathbf{\hat{\epsilon}}^{i} = w * \mathbf{c}^{i} + (1 - w) * \mathbf{\epsilon}_{\text{shared}}, \label{eqn:cond_noise} \\
    \mathbf{\hat{z}}^{i}_T = \sqrt{\bar{\alpha}_T} \mathbf{\hat{\epsilon}}^{i} + \sqrt{1 - \bar{\alpha}_T} \mathbf{\epsilon}^{i}, \label{eqn:noise_init}
\end{align}

\noindent where $\mathbf{c}^{i}$ represents the low-frequency coordinate/depth-map information for view $i$, which is linearly combined with the shared noise $\mathbf{\epsilon}_{\text{shared}}$ using weight $w$. For depth conditioning, we set $\mathbf{c}^{i}$ as the normalized depth maps per view, whereas for panoramic images we use the normalized coordinates for each view transformed into the coordinate space of the center view (see supplemental material for further detail). We refer to this combination of coordinate-based information and shared noise as \emph{coordinate noise}. When $w = 0$, there is no low-frequency condition -- a setting we refer to as \emph{simple shared noise}. The updated latent $\mathbf{\hat{z}}_T$ is then used in place of $\mathbf{z}_{T}$ when initializing noise for image generation.

\subsection{Fourier-based Attention Module}
\label{sec:fba}
Based on the insights from works \cite{ge2023preserve,wu2023freeinit,ren2024consisti2v} incorporating shared noise or low spatial frequency information when generating multi-view images, we hypothesize that attending to coordinate noise features -- particularly in non-overlapping regions of multi-view images -- will improve the consistency in global appearance across a generated scene. Building on MVDiffusion's \cite{tang2023mvdiffusion} \emph{correspondence-aware attention} (CAA) modules, we propose a Fourier-based attention (FBA) module that incorporates coordinate noise features with spatial frequencies selected dependent on the denoising time step. Similar to the CAA blocks, each FBA block contains the attention module and a residual network with zero-initialized convolution layers. Whereas the CAA modules are intended to attend to corresponding points in overlapping image regions, the intuition of our approach is to inject shared noise to better align the overall scene appearance in the non-overlapping image regions.

Specifically, for a given time step $t$ and noise initialization (\ie Equation \ref{eqn:noise_init}), let $\mathbf{F}^{i}_{t}$ be the feature maps of the U-Net denoising network for the view indexed by $i \in \left[0, N - 1 \right]$. Let $\mathbf{G}^{i}_{t}$ then be the features obtained when using the \emph{coordinate noise} $\mathbf{\hat{\epsilon}}$ (\ie Equation \ref{eqn:cond_noise}) to set $\mathbf{z}_{t}$ in Equation \ref{eqn:forward_noise_step}. Since these feature maps are the target of the attention module, we gather $\mathbf{G}^{i}_{t}$ from each preceding layer without applying the FBA blocks. We then apply the Fast Fourier Transform (FFT; Equation \ref{eqn:fft}) to obtain the spatial frequency domain representations of features $\mathbf{G}^{i}_{t}$, allowing us to select and attend to higher-frequency features to align appearance, where lower frequencies more represent image structure \cite{wu2023freeinit}. The inverse transform $\mathcal{F}^{-1}$ is similarly applied to transform back to the image domain (\ie FFT and its inverse are applied to the height and width dimensions).

\begin{align}
    \mathcal{F}(m, n) &= \sum_{h, w}\mathbf{x}(h, w)\exp{-j 2 \pi \left(\frac{h}{H} m + \frac{w}{W} n \right)}, \label{eqn:fft} \\
    j^2 &= -1 \nonumber
\end{align}

To implement this method, we combine the \emph{correspondence-aware attention} within overlapping regions and our \emph{Fourier-based attention} in non-overlapping regions. Using the known homography matrices relating each view, we can obtain the mask of overlapping regions $\mathbf{M}^{i,j}_{\text{ovr}}$ between views $i$ and $j$. Formally, for the set of source feature maps $\mathbf{F}^{i}_{t}$, we select the corresponding features $\mathbf{\bar{F}}^{j}_{t}$ in overlapping regions of each target view $j$ and the spatially-filtered features $\mathbf{\bar{G}}^{j}_{t}$ in non-overlapping regions. Following \cite{tang2023mvdiffusion}, the features in overlapping regions of target views are interpolated from the target coordinate space $v$ to map onto the corresponding locations in the source coordinate space $u$ (see \cite{tang2023mvdiffusion} for further detail):

\begin{equation}
    \mathbf{\bar{F}}^{j}_{t}(v^{j}) = \mathbf{F}^{j}_{t}(v^{j}) + \gamma(u^{j}_{*} - u),
    \label{eqn:caa}
\end{equation}

\noindent where $\gamma(\cdot)$ represents the positional encoding of the displacement between corresponding coordinates $u^{j}_{*}$ and original coordinates $u$ in the source view.

The \emph{coordinate noise} features $\mathbf{G}^{j}_{t}$ are spatially filtered by masking portions of the frequency spectrum dependent on the time step. Specifically, for each time step we select spatial frequencies proportional to the noise level by creating a mask with ones everywhere except within a central region whose radius is dependent on the time step:

\begin{align}
    r_{t} &= 1 - \frac{t}{T}, \\
    \mathbf{M}^{r_{t}}_{\mathcal{F}} &= \left( 1 - \mathbbm{1}_{(h, w) \in [-r_{t} H: r_{t} H, - r_{t} W: r_{t} W]} \right),
    \label{eqn:fft_mask}
\end{align}

\noindent where $H$ and $W$ are the height and width of the U-Net features, respectively. The mask $\mathbf{M}^{r_{t}}_{\mathcal{F}}$ therefore represents the spatial frequencies corresponding to the \emph{coordinate noise} features $\mathbf{G}^{j}_{t}$. The mask is then element-wise multiplied with the frequency signal to obtain the non-overlapping features to be attended, as shown in Equation \ref{eqn:fft_feat} below.

\begin{equation}
    \mathbf{\bar{G}}^{j}_{t} = \mathcal{F}^{-1}(\mathbf{M}^{r_{t}}_{\mathcal{F}} \odot \mathcal{F}(\mathbf{G}^{j}_{t})) + \gamma(1 - r_{t}),
    \label{eqn:fft_feat}
\end{equation}

\noindent where $\gamma(1 - r_{t})$ represents the positional encoding of the time step-dependent radius. Finally, the target features of overlapping and non-overlapping regions are combined using the mask of overlapping regions $\mathbf{M}^{i,j}_{\text{ovr}}$:

\begin{equation}
    \mathbf{V}^{i,j}_{t} = \mathbf{M}^{i,j}_{\text{ovr}} \odot \mathbf{\bar{F}}^{j}_{t} + (1 - \mathbf{M}^{i,j}_{\text{ovr}}) \odot \mathbf{\bar{G}}^{j}_{t}.
    \label{eqn:tgt_feat}
\end{equation}

We then apply attention from each query source view $\mathbf{F}^{i}$ to the set of target views $\mathbf{V}^{i}$:

\begin{equation}
    \mathbf{\hat{F}}^{i} = \texttt{SoftMax}\left([\mathbf{W}_{Q} \mathbf{F}^{i}] \cdot [\mathbf{W}_{K} \mathbf{V}^{i}]\right) \mathbf{W}_{V} \mathbf{V}^{i},
    \label{eqn:attn}
\end{equation}

\noindent where $\mathbf{W}_{Q}$, $\mathbf{W}_{K}$, and $\mathbf{W}_{V}$ are the weights corresponding to the query, key, and value commonly used in attention modules \cite{vaswani2017attention}.

\subsection{Prompt Cross Attention Loss}
\label{sec:xa_loss}
In order to improve the spatial consistency of features across views, we propose a novel loss that ensures that the cross attention maps between each prompt and view are consistent with those in the ground truth scene. This method takes inspiration from the cross attention loss proposed in \cite{parmar2023zero}, which was shown to improve structural consistency during image-to-image editing with diffusion models. We extend this cross attention loss to multi-view image generation by
computing it on the attention between each view's prompt and all other view features. This ensures that the prompt-based attention between disparate views is consistent with the ground truth scene.

We implement the cross attention loss at each $16 \times 16$ resolution cross attention module. In order to obtain the ground truth attention maps $\mathcal{M}^{l}_{t}$, we first pass the clean latent views $\mathbf{z}^{1:n}_0$ through the U-Net to collect the noise-free attention maps. The cross attention loss is then computed at each applicable layer $l$ as follows:

\begin{equation}
    \mathcal{L}^{l}_{XA} = \lVert \mathcal{M}^{l}_t - \mathcal{M}^{l}_0 \rVert.
    \label{eqn:xa_loss}
\end{equation}

\subsection{Training Paradigm}
In order to train our FBA blocks, we start from a diffusion model trained on single views. For depth-to-image training, the diffusion model is first fine-tuned to generate images at the $192 \times 256$ image resolution. This training is done using single-view images only. During training of the FBA blocks, we randomly select a sequence of $n$ partially overlapping views from the dataset and a single time step for all views $t \sim \mathcal{U}[1, T]$. During this stage, we keep the original U-Net model parameters frozen and train the proposed FBA blocks end-to-end to minimize the following overall loss function:

\begin{equation}
    \mathcal{L} = \mathcal{L}_{LDM} + \lambda \sum_{l \in L}{\mathcal{L}^{l}_{XA}},
    \label{eqn:total_loss}
\end{equation}

\noindent where $L$ denotes the set of layer indices corresponding to attention maps processing $16 \times 16$ spatial resolution and $\lambda$ is set to $10$.

\section{Experiments}
\label{sec:experiments}
\begin{figure}
    \centering
    \includegraphics[width=0.98\linewidth]{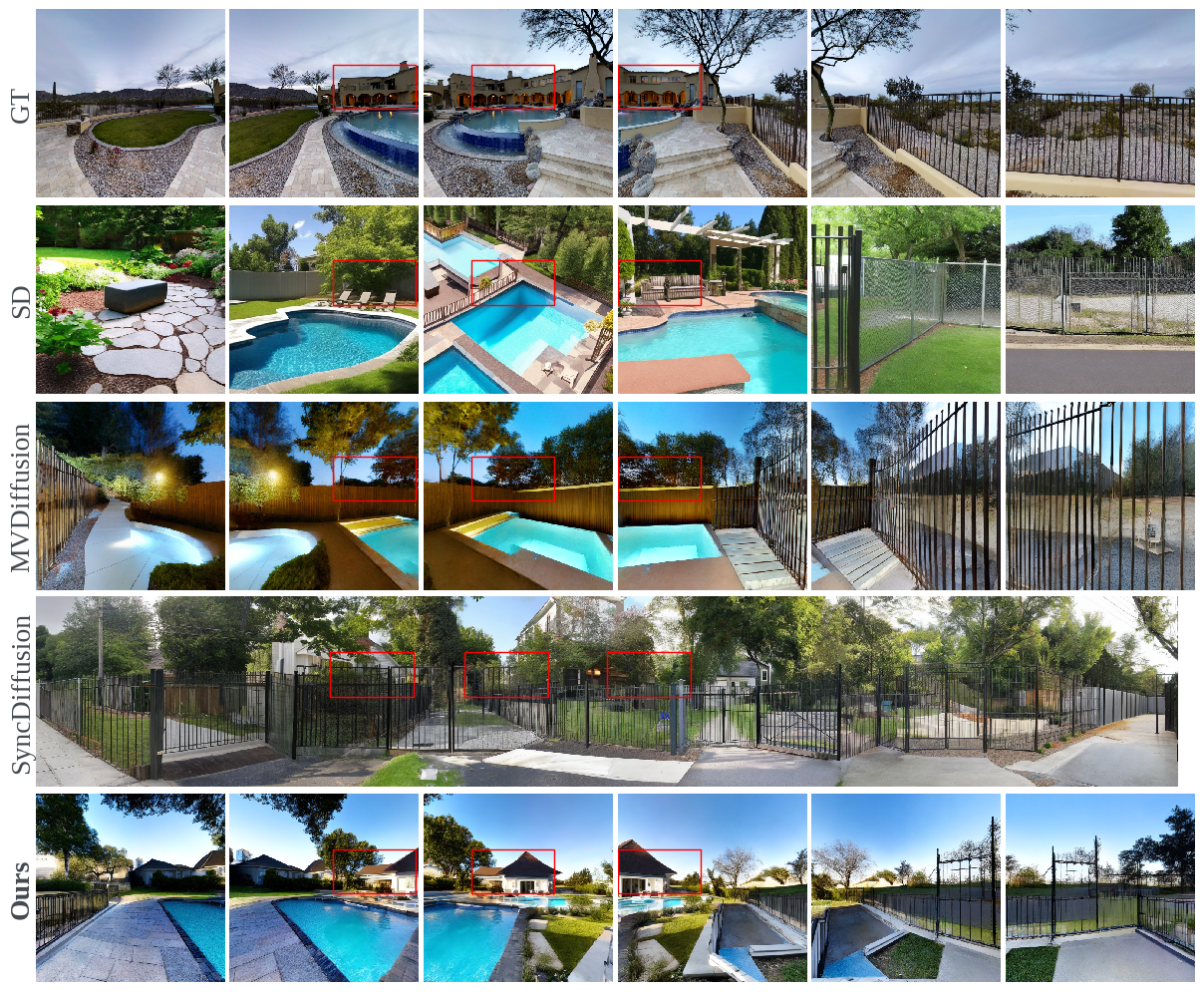}
    \caption{Qualitative comparison for panoramic image generation. Colored boxes highlight misalignment with prompt ``a house with a pool in the backyard''. See Section \ref{sec:qual_eval} for further detail.}
    \label{fig:pano_qual}
\end{figure}

\begin{table}
    \centering
    \begin{tabular}{c|c|c}
        \hline
        Method & FID $\downarrow$ & CLIP Score $\uparrow$ \\
        \hline
        Baseline LDM \cite{rombach2022LDM} & 45.6 & \textcolor{gray}{25.6} \\
        MVDiffusion \cite{tang2023mvdiffusion} & 30.3 & 24.3 \\
        SyncDiffusion \cite{lee2023syncdiffusion} & 51.4 & 20.0 \\
        Ours & \textbf{22.36} & \textbf{24.7} \\
    \end{tabular}
     \caption{Quantitative evaluation of image quality in panoramic experiments compared with baseline.}
    \label{tab:pano_eval_iq}
\end{table}

\begin{table}
    \begin{tabular}{c|c|c|c}
    \hline
    Method & PSNR $\uparrow$ & Ratio $\uparrow$ & I-LPIPS $\downarrow$ \\
    \hline
    GT & 37.7 & 1.00 & 0.71 \\
    \hline
    Baseline LDM \cite{rombach2022LDM} & 9.10 & 0.24 & 0.80 \\
    MVDiffusion \cite{tang2023mvdiffusion} & 22.2 & 0.60 & 0.79 \\
    SyncDiffusion \cite{lee2023syncdiffusion} & - & - & \textbf{0.64} \\
    Ours & \textbf{24.7} & \textbf{0.66} & 0.75 \\
    \end{tabular}
        \caption{Quantitative evaluation of multi-view consistency. Bold text indicates best performance among methods addressing multi-view consistency.}
    \label{tab:pano_eval_mv}
    \vspace{-\baselineskip}
\end{table}

\begin{figure*}
    \centering
    \includegraphics[width=0.98\linewidth]{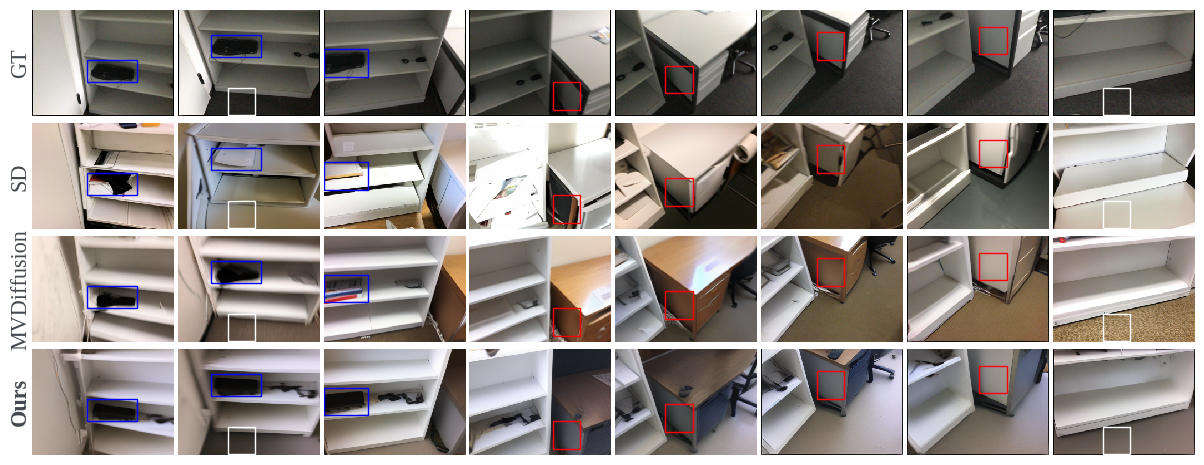}
    \caption{Qualitative comparison for multi-view depth-to-image generation. Colored boxes highlight inconsistencies in baselines relative to our method. Blue, red and white boxes demonstrate how small objects, large objects and environment (\ie non-overlapping regions) resp. change appearance in our baselines. We qualitatively outperform our baselines. See Section \ref{sec:qual_eval} for further detail.}
    \label{fig:depth_qual}
\end{figure*}

We evaluate our method in two settings: multi-view panoramic and depth-conditioned image generation. To evaluate panoramic image generation, we use the Matterport3D\footnote{https://matterport.com/legal/matterport-end-user-license-agreement-academic-use-model-data} dataset~\cite{chang2017matterport3d} consisting of 10,912 panoramic indoor scenes. Following \cite{tang2023mvdiffusion}, we separate the dataset into 9820 panoramic sequences used for training and 1092 for evaluation. To evaluate performance of multi-view depth-to-image generation, we use the ScanNet dataset~\cite{dai2017scannet}, containing 1513 scenes for training and 100 scenes for evaluation. We discuss method implementation details in Section~\ref{sec:impl}, baselines in Section~\ref{sec:baselines}, quantitative, qualitative and ablation results in Section~\ref{sec:quantitative}, Section~\ref{sec:qual_eval} and Section~\ref{sec:ablation} respectively. 
Additional results and implementation details can be found in the Supplementary Materials.

\begin{table*}
  \centering
   
    \begin{adjustbox}{max width=\textwidth}
    \begin{tabular}{c|c|c}
        \hline
        Method & FID $\downarrow$ & CLIP Score $\uparrow$ \\
        \hline
        ControlNet \cite{zhang2023adding} & 38.3 & 20.5 \\
        MVDiffusion \cite{tang2023mvdiffusion} & \textbf{23.7} & \textbf{24.3}  \\
        Ours & 27.0 & 23.8 \\
    \end{tabular}
    \begin{tabular}{c|c|c|c}
        \hline
        Method & PSNR $\uparrow$ & Ratio $\uparrow$ & Intra-LPIPS $\downarrow$ \\
        \hline
        GT & 15.4 & 1.00 & 0.58\\
        \hline
        ControlNet \cite{zhang2023adding} & 9.05 & 0.62 & 0.70 \\
        MVDiffusion \cite{tang2023mvdiffusion} & 13.0 & 0.87 & 0.67 \\
        Ours & \textbf{13.9} & \textbf{0.94} & \textbf{0.64} \\
    \end{tabular}
        \end{adjustbox}
         \caption{Quantitative comparison with baseline methods for depth-conditioned experiments. Left: image quality, right: multi-view consistency. Bold text indicates best performance.}
    \label{tab:depth_mv}
    \vspace{-\baselineskip}
\end{table*}

\subsection{Implementation Details}
\label{sec:impl}
We implement our method with PyTorch using the latent diffusion model architecture provided from Diffusers \cite{von-platen-etal-2022-diffusers}. During training, we freeze the parameters of the denoising U-Net and train only our newly added modules. We train each method using 4 nodes with 8 x A100 GPUs each for 20 epochs in the depth-to-image experiment and 10 epochs in the panoramic experiment. We use a per-GPU batch size of 1 and a learning rate $1e^{-4}$ and $2e^{-4}$ for the depth-conditioned and panoramic experiments, respectively.

During inference, we generate 8 views simultaneously for both depth-to-image and panoramic experiments. For panoramic image generation, each view is separated by a rotation angle of 45 degrees. For depth-to-image generation, we follow the method described in \cite{tang2023mvdiffusion} for generating key frames and interpolation for denser image generation. The key-frame views are curated to maintain approximately 65\% overlap between each pair of key frames. For interpolating between views, the generated key frames are used to condition the generation of the interpolated views as described in \cite{tang2023mvdiffusion}.

\subsubsection{Evaluation Metrics}
We utilize multiple metrics to evaluate image quality and multi-view consistency. To evaluate the image quality of generated multi-view scenes, we compute the following metrics:

\begin{itemize}
    \item Frechet Inception Distance (FID) \cite{heusel2017gans}: measures the distribution gap between generated and real images.
    \item CLIP Score (CS) \cite{radford2021learning}: measures the text and generated image similarity using the CLIP model.
\end{itemize}

In order to evaluate consistency of generated multi-view images, we use the following metrics:

\begin{itemize}
    \item Overlap Peak Signal-to-Noise Ratio (PSNR) \cite{tang2023mvdiffusion}: PSNR between all overlapping regions, compared as a ratio between generated and real images.
    \item Intra-LPIPS \cite{zhang2018unreasonable}: measures the coherence of panoramic images, computed as the average LPIPS distance of all combinations of generated image pairs for a scene. 
\end{itemize}

We use the same evaluation method for each experiment as described in \cite{tang2023mvdiffusion}. In brief, to evaluate multi-view consistency we compute overlapping PSNR ratios between consecutive generated images relative to the ground truth comparisons. 

\subsection{Baselines}
\label{sec:baselines}
We evaluate our performance against the following baseline methods for the panoramic experiment:

\begin{itemize}
    \item MVDiffusion~\cite{tang2023mvdiffusion} uses the \emph{correspondence-aware} attention module to attend to a nearby set of views for panoramic experiments.
    \item Baseline LDM~\cite{rombach2022LDM} constitutes the baseline pre-trained model upon which MVDiffusion is trained.
    \item SyncDiffusion~\cite{lee2023syncdiffusion} is a training-free method designed for panoramic image generation with diffusion models.
\end{itemize}

To evaluate performance for the depth-to-image experiment, we compare against the following baselines:

\begin{itemize}
    \item MVDiffusion~\cite{tang2023mvdiffusion} incorporates a \emph{correspondence-aware} attention module (CAA) that attends to nearby views with corresponding points. As described above, we utilize this CAA mechanism within our own attention module.
    \item ControlNet~\cite{zhang2023adding} is a popular method for conditioning LDMs, which in our experiments can be used for evaluating depth-to-image generation.
\end{itemize}

Whereas our and other baseline methods generate $n$ views in parallel, SyncDiffusion \cite{lee2023syncdiffusion} generates a single panoramic image with $512 \times 3072$ resolution using a single text prompt. In order to compare against the other baselines, we first combine the per-view text prompts used in our method into a single prompt describing the full scene. Then following image generation, we split the panoramic image into six non-overlapping views with resolution $512 \times 512$ as described in \cite{lee2023syncdiffusion}. We use these views for quantitative evaluations of FID, CLIP Score, and Intra-LPIPS.

\subsection{Quantitative Evaluation}
\label{sec:quantitative}
\subsubsection{Panoramic Experiment}
We report quantitative results for the panoramic image generation experiment in Tables \ref{tab:pano_eval_iq} and \ref{tab:pano_eval_mv}. As shown in the tables, our method consistently outperforms the baselines across most metrics, particularly FID and overlapping PSNR. Although SyncDiffusion achieves a lower Intra-LPIPS, the value is far lower even than the ground truth images ($0.64$ vs. $0.71$). This is likely due to the fact that their method aims to increase coherence across all views, resulting in similar content repeated across the scene (\eg, see Figure \ref{fig:pano_qual}). This is also supported by their relatively low CLIP Score ($20.0$ \vs $24.7$ in our method), which indicates that their generated images do not respect the provided text prompts. Meanwhile, among methods that aim to improve multi-view consistency, ours has the highest CLIP Score.

\subsubsection{Depth-to-Image Experiment} We report quantitative results for the depth-to-image experiment in Table \ref{tab:depth_mv}. As seen in the table, our method greatly improves the multi-view consistency compared to MVDiffusion ($0.94$ \vs $0.87$ ratio and $0.64$ \vs $0.67$). In addition to non-overlapping improvements from FBA blocks, we attribute improvements in overlapping regions to two primary differences: 1) coordinate-based noise initialization better informs scene structure and 2) the prompt cross attention loss improves prompt-spatial alignment across views. However, we do observe slightly lower performance in the depth-to-image experiment in terms of FID, while CLIP Score demonstrates competitive performance, which we discuss in further detail in the supplemental material.

\subsection{Qualitative Evaluation}
\label{sec:qual_eval}
\subsubsection{Panoramic Experiment}
We show qualitative results in Figure \ref{fig:pano_qual} for the panoramic experiment. When compared with MVDiffusion and SyncDiffusion, we observe several instances where generated views are missing attributes from the provided prompt. In this case, the central views were conditioned on the prompt ``a house with a pool in the backyard'', but the generated scene from MVDiffusion's method does not contain a house. SyncDiffusion's method generates a house but fails to generate the pool. We hypothesize our method achieves better prompt alignment in such cases due to the XA loss, which trains the model to generate images that preserve the attention maps between each prompt and all views. 

\subsubsection{Depth-to-Image Experiment}
\begin{figure}
  \centering
  \includegraphics[width=\linewidth]{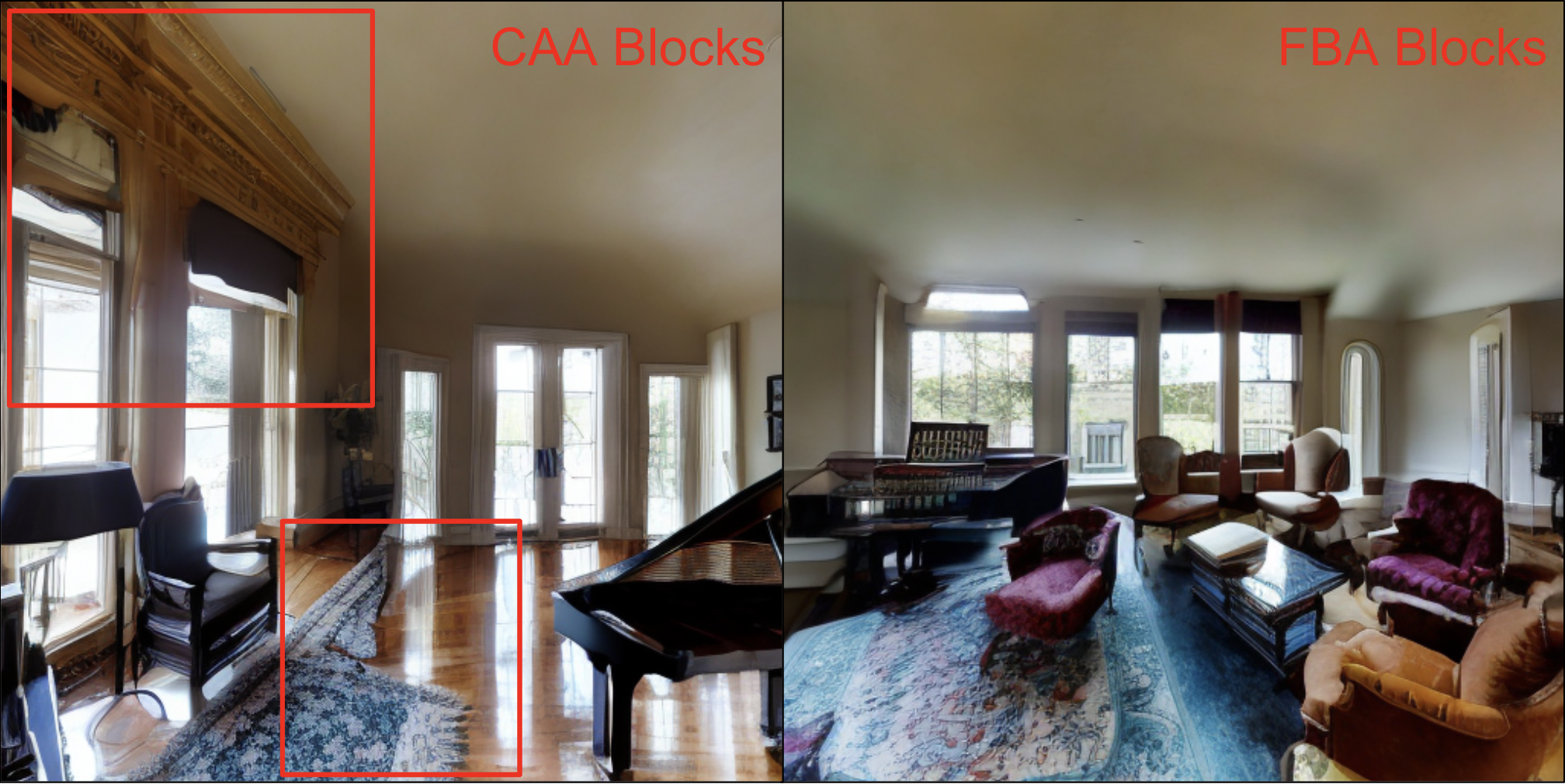}
  \caption{Color/texture inconsistencies using CAA (MVDiffusion) \vs FBA (ours) blocks.}
   \label{fig:fba_ablation}
\end{figure}

As shown in Figure \ref{fig:depth_qual}, the other baseline methods exhibit multiple inconsistencies across generated views. Specifically, the blue box in the first three columns demonstrate how small objects may change appearance when generating with SD or MVDiffusion. The red box in the middle columns highlight changes in texture color for larger objects, where MVDiffusion the desk color changes from brown to white. Finally, the white box in columns 2 and 8 showcase our method's ability to generate images with a globally consistent appearance, whereas in MVDiffusion's scene the floor texture changes across the scene. We argue that the incorporation of our FBA blocks, which attend to the non-overlapping regions, help in achieving higher consistency across more disparate views of a scene. This is supported by Figure \ref{fig:fba_ablation}, which demonstrates that when using MVDiffusion's CAA blocks \vs our FBA blocks abrupt changes in colors and textures are observed.

\subsection{Ablation Studies}
\label{sec:ablation}
\begin{table}[t]
    \centering
    \begin{tabular}{l|c|c|c|c|c}
        \hline
        Method & FID & CS & PSNR & Ratio & I-LPIPS\\
        \hline
        Shared Noise & 22.1 & 24.7 & 23.6 & 0.63 & 0.79 \\
        Coord. Noise & \textbf{19.5} & \textbf{24.9} & 24.2 & 0.65 & 0.78 \\
        FBA Blocks & 21.0 & 24.9 & 23.9 & 0.64 & 0.78 \\
        Full Model & 22.4 & 24.7 & \textbf{24.7} & \textbf{0.66} & \textbf{0.75} \\
    \end{tabular}
     \caption{Ablation of FBA Blocks.}
    \label{tab:pano_ablation}
    \vspace{-\baselineskip}
\end{table}

In Table \ref{tab:pano_ablation}, we report quantitative results of an ablation study for the panoramic image generation experiment (see supplemental material for qualitative comparisons). We start by evaluating performance when only using \emph{Shared Noise} (Eqn. \ref{eqn:shared_noise_init}), which greatly improves FID relative to the MVDiffusion baseline, but provides more modest improvements in multi-view consistency. Using instead the \emph{Coordinate Noise} (Eqn. \ref{eqn:cond_noise}) for initialization provides further improvements in both image quality and multi-view consistency. We observe similar performance when introducing the FBA blocks (Sec. \ref{sec:fba}); however, when combined with the cross-attention loss (Sec. \ref{sec:xa_loss}), we observe our best overall performance. Figure \ref{fig:fba_ablation} further shows the qualitative improvements when using FBA \vs CAA blocks.
\section{Conclusion}
\label{sec:conclusion}

In this paper we address the challenge of multi-view consistent text-to-image generation. We propose a diffusion model that utilizes the Fourier space to select features for attention in non-overlapping regions. We further propose a novel noise initialization technique and cross-attention that ensure higher multi-view consistency in the overlapping regions. As shown qualitatively and quantitatively we outperform SOTA baselines and achieve multi-view consistency while maintaining the diversity in the generated images. In the future, we want to extend this work to generate high-fidelity, multi-view and temporally-consistent videos from prompts, conditioned on depth-maps.
\beginsupplement

\section{Noise Initialization}
\subsection{Implementation Details}
In this section we provide further implementation details of our coordinate-based noise initialization. For each set of multi-view images, we first sample a ``shared noise'' that is used across all views (\ie $\mathbf{\epsilon}_{\text{shared}}$ in Eqn. \ref{eqn:shared_noise_init}). To provide the model with low spatial frequency information related to the change in camera pose across views, we transform normalized pixel coordinates from each view into the space of the center view. We then take the cosine of these values to remap pixel coordinates into the range $[-1, 1]$. These transformed pixel coordinates are then combined with the shared noise according to Eqn. \ref{eqn:cond_noise}. The coordinate noise for each view $\mathbf{\hat{\epsilon}}^{i}$ is then combined with per-view independent noise $\mathbf{\epsilon}^{i}$ as shown in Eqn. \ref{eqn:noise_init}.

\subsection{Quantitative Comparisons of Noise Initialization Methods}
In order to further evaluate the choice of coordinate noise, we compare against other relevant methods for incorporating shared noise or low-frequency information (Table \ref{tab:noise_comp}). The first comparison of interest is ``mixed noise'' \cite{ge2023preserve}, which uses a combination of shared noise across views and independent noise per view. This is similar to our ``shared noise'' condition in our ablation study in the main paper (Table \ref{tab:pano_ablation}) but uses a different weighting scheme (Eqn. \ref{eqn:mixed_noise} with $\alpha = 1$). As shown in Table \ref{tab:noise_comp}, our shared noise implementation provides better performance across all metrics except Intra-LPIPS (compare first two rows). 

\begin{equation}
    \mathbf{\epsilon}^{i}_{\text{mixed}} = \mathbf{\epsilon}_{\text{shared}} \frac{\alpha^2}{1 + \alpha^2} + \mathbf{\epsilon}^{i} \frac{1}{1 + \alpha^2}
    \label{eqn:mixed_noise}
\end{equation}

Next, we compare the effect of using our ``coordinate noise'' implementation \vs combining low-frequency coordinate noise and high-frequency independent noise, which has been suggested in recent work conditioning on images \cite{wu2023freeinit,ren2024consisti2v}. Although we do not condition directly on image frames, it's clear that the combination of low-frequency coordinate noise and high-frequency independent noise is not as effective as our implementation using Eqn. \ref{eqn:noise_init} (compare last two rows of Table \ref{tab:noise_comp}).

Overall, it is interesting to note that although our coordinate noise method provides substantial improvements in FID and overlapping PSNR, mixed noise obtains better performance when measuring Intra-LPIPS. 

\begin{table}[h]
    \caption{Comparison of noise initialization methods in the panoramic experiment.}
    \centering
    \begin{tabular}{c|c|c|c|c|c}
        \hline
        Method & FID $\downarrow$ & CLIP Score $\uparrow$ & PSNR $\uparrow$ & Ratio $\uparrow$ & Intra-LPIPS $\downarrow$ \\
        \hline
        Mixed Noise \cite{ge2023preserve} & 23.25 & 24.69 & 23.25 & 0.624 & \textbf{0.719} \\
        Shared Noise (Eqn. \ref{eqn:shared_noise_init}) & 22.06 & 24.71 & 23.63 & 0.635 & 0.794 \\
        Low Freq. Coord. Noise & 36.99 & 23.14 & 21.63 & 0.582 & 0.777 \\
        Coord. Noise (Eqn. \ref{eqn:cond_noise}) & \textbf{19.55} & \textbf{24.95} & \textbf{24.25} & \textbf{0.651} & 0.776 \\
        \hline
    \end{tabular}
    \label{tab:noise_comp}
\end{table}

\begin{figure}[h]
  \centering
  \includegraphics[width=0.98\linewidth]{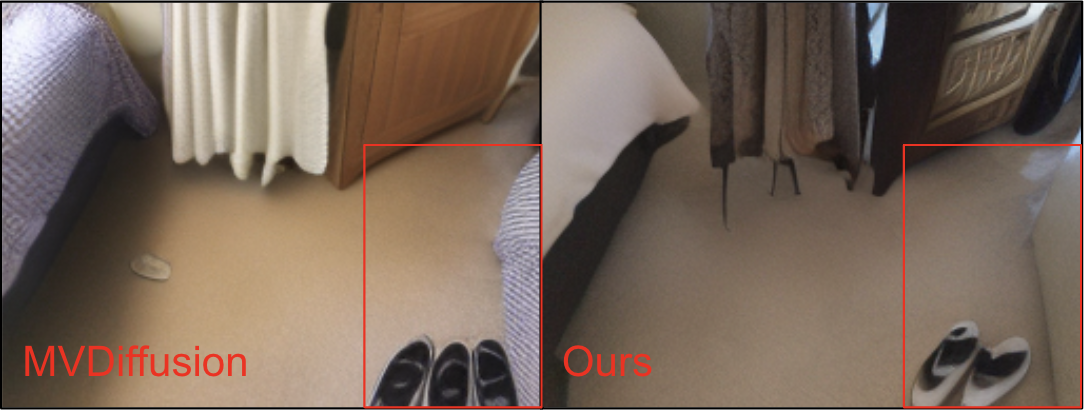}
   \caption{MVDiffusion \vs our method with an imprecise prompt ``\textit{a pair of shoes sitting on the floor next to a bed}.''}
   \label{fig:prompt_errors}
\end{figure}

\section{FID/CLIP Score Differences Between Experiments}
As noted in the main paper, we observed improved performance as measured by FID and CLIP Score compared to MVDiffusion in the panoramic but not the depth-to-image experiment (\cf Tables 1 \& 3). One explanation for this performance difference is that ScanNet text prompts provided by \cite{tang2023mvdiffusion} using blip2 were often imprecise or inconsistent across views. Since MVDiffusion's method does not account for non-overlapping regions, their method is susceptible to issues like that shown in Figure \ref{fig:prompt_errors} for imprecise prompts (here, the prompt ``a pair of shoes sitting on the floor next to a bed'' leads to hallucinations of a second bed). These errors can lead to better CLIP Score performance at the expense of multi-view consistency. Furthermore, inconsistent prompts across a scene could negatively impact FID for our method compared with MVDiffusion, which may exhibit errors only in single views without reconciling across a scene.

\section{Additional Ablation Studies}

In order to further evaluate our design choices for noise initialization, we compare results from experiments varying the weight parameter $w$ from Eqn. \ref{eqn:cond_noise}. The results shown in Table \ref{tab:ablate_noisew} indicate that setting the weight $w = 0.5$ indeed provides the optimal result. However, it is interesting to note that this paramter appears to primarily affect FID and overlapping PSNR metrics. For these metrics, performance is noticeably -- albeit not substantially -- worse in either direction away from $0.5$.

\begin{table}[h]
    \caption{Ablation of weight parameter $w$ in Eqn. \ref{eqn:cond_noise} in the panoramic experiment.}
    \centering
    \begin{tabular}{c|c|c|c|c|c}
        \hline
        Method & FID $\downarrow$ & CLIP Score $\uparrow$ & PSNR $\uparrow$ & Ratio $\uparrow$ & Intra-LPIPS $\downarrow$ \\
        \hline
        Shared Noise ($w = 0.0$) & 22.06 & 24.71 & 23.63 & 0.635 & 0.794 \\
        Coord. Noise ($w = 0.25$) & 19.71 & 24.90 & 23.92 & 0.643 & 0.779 \\
        Coord. Noise ($w = 0.5$) & \textbf{19.55} & \textbf{24.95} & \textbf{24.25} & \textbf{0.651} & \textbf{0.776} \\
        Coord. Noise ($w = 0.75$) &  21.02 & 24.90 & 23.59 & 0.634 & 0.787 \\
        Coord. Noise ($w = 1.0$) & 21.70 & 24.90 & 23.91 & 0.643 & 0.781 \\
        \hline
    \end{tabular}
    \label{tab:ablate_noisew}
\end{table}

We additionally compare performance when using a binary high pass filter (HPF) mask (Eqn. \ref{eqn:fft_mask}) \vs a Gaussian HPF approach as well as when using a time-dependent (HPF-$r_{t}$) \vs constant low pass stop frequency (LPF-$0.25$, using stop frequency from \cite{wu2023freeinit,ren2024consisti2v}). The results shown in Table \ref{tab:ablate_fftmask} demonstrate that there is minimal difference between the binary or Gaussian HPF mask. However, we observe that using a time-dependent HPF mask provides substantially better performance.

\begin{table}[h]
    \centering
    \begin{threeparttable}
        \caption{Comparison of binary and Gaussian high (HPF) or low (LPF) pass filters (Eqn. \ref{eqn:fft_mask}) in the panoramic experiment.}
        \begin{tabular}{c|c|c|c|c|c}
            \hline
            Method & FID $\downarrow$ & CLIP Score $\uparrow$ & PSNR $\uparrow$ & Ratio $\uparrow$ & Intra-LPIPS $\downarrow$ \\
            \hline
            Gaussian LPF-0.25 mask & 23.99 & 24.71 & 23.13 & 0.621 & 0.771 \\
            Gaussian HPF-$r_{t}$ mask & 22.59 & \textbf{24.84} & 24.47 & 0.657 & 0.762 \\
            Binary HPF-$r_{t}$ mask (Eqn. \ref{eqn:fft_mask}) & \textbf{22.36} & 24.68 & \textbf{24.67} & \textbf{0.662} & \textbf{0.755} \\
            \hline
        \end{tabular}
        \begin{tablenotes}
          \small
          \item  \textit{Note}: Filters are either time-dependent (\ie ``HPF-$r_{t}$'' where $r_{t}$ is the radius defined in Eqn. \ref{eqn:fft_mask}) or use a normalized stop frequency of 0.25 (\ie ``LPF-$0.25$'').
        \end{tablenotes}
        \label{tab:ablate_fftmask}
    \end{threeparttable}
\end{table}

Finally, we further validate the design choice of our time-dependent Fourier-based attention module. Specifically, we consider the following conditions: no spatial frequency filtering (``No filter''), time-dependent low pass filtering (``LPF-$r_{t}$''), as well as low and high pass filtering using the inverse relationship with denoising time steps (``LPF-$(1 - r_{t})$'' and ``HPF-$(1 - r_{t})$'', respectively).  In the latter two conditions, the radius $r_{t}$ of the spatial frequency mask in Eqn. \ref{eqn:fft_mask} decreases from $1$ to $0$ across denoising time steps. For low pass filtering (\ie ``LPF-$(1 - r_{t})$''), this means that all frequencies are included in $\mathbf{\bar{G}}^{j}_{t}$ (Eqn. \ref{eqn:fft_feat}) at the noisiest time steps and only the lowest frequencies are included at the least noisy time steps.

As shown in Table \ref{tab:ablate_filter}, our method of selecting the full spectrum of spatial frequencies for attention at noisier time steps and high spatial frequencies at less noisy time steps (\ie ``HPF-$r_{t}$'') provides the best overall performance, particularly for FID and overlapping PSNR. Similar to our ablation of the weight parameter in Eqn. \ref{eqn:cond_noise} (Table \ref{tab:ablate_noisew}), we observe relatively less variation across conditions for the CLIP Score and Intra-LPIPS metrics.

\begin{table}[h]
    \centering
    \begin{threeparttable}
        \caption{Comparison of time-dependent low or high pass filters in the panoramic experiment.}
        \begin{tabular}{c|c|c|c|c|c}
            \hline
            Method & FID $\downarrow$ & CLIP Score $\uparrow$ & PSNR $\uparrow$ & Ratio $\uparrow$ & Intra-LPIPS $\downarrow$ \\
            \hline
            No filter & 25.89 & \textbf{24.85} & 23.31 & 0.626 & 0.747 \\
            LPF-$(1 - r_{t})$ & 29.71 & 24.78 & 22.66 & 0.609 & 0.768 \\
            LPF-$r_{t}$ & 23.81 & 24.75 & 24.12 & 0.648 & \textbf{0.740} \\
            HPF-$(1 - r_{t})$ & 23.57 & 24.57 & 24.00 & 0.645 & 0.772 \\
            HPF-$r_{t}$ (Eqn. \ref{eqn:fft_mask}) & \textbf{22.36} & 24.68 & \textbf{24.67} & \textbf{0.662} & 0.755 \\
            \hline
        \end{tabular}
        \begin{tablenotes}
          \small
          \item  \textit{Note}: The low pass filter (LPF) is defined as $1 - \mathbf{M}^{r_{t}}_{\mathcal{F}}$ and, \eg, ``HPF-$(1 - r_{t})$'' implies $\mathbf{M}^{(1 - r_{t})}_{\mathcal{F}}$.
        \end{tablenotes}
        \label{tab:ablate_filter}
    \end{threeparttable}
\end{table}

\section{Additional Qualitative Examples}
In this section, we provide further qualitative examples of our method in comparison to baselines in the depth-to-image and panoramic image generation experiments.

\begin{figure}[h]
    \centering
    \includegraphics[width=0.98\linewidth]{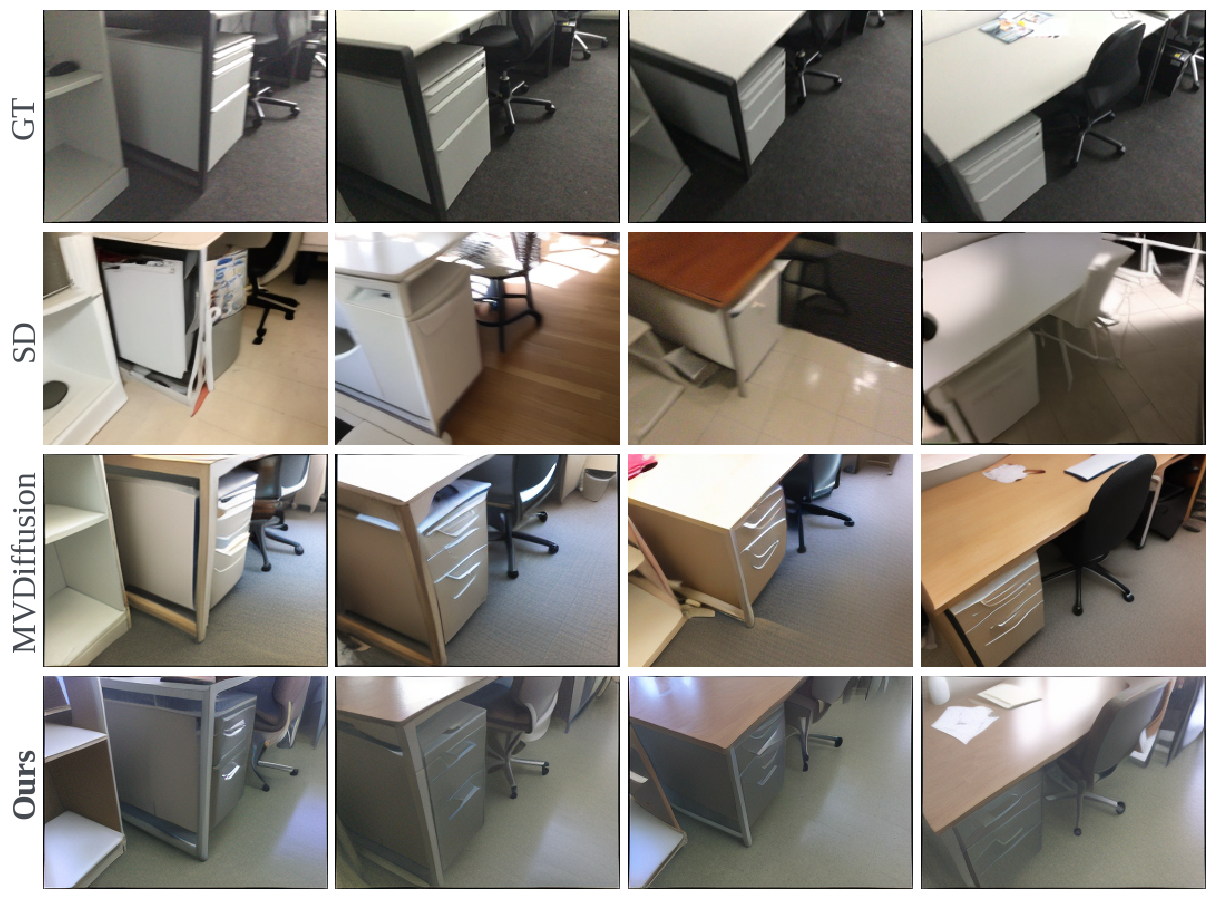}
    \caption{Depth-to-image generation using the prompt ``\textit{a desk with a chair and a filing cabinet}.''}
    \label{fig:depth_qual1}
\end{figure}

\begin{figure}[h]
    \centering
    \includegraphics[width=0.98\linewidth]{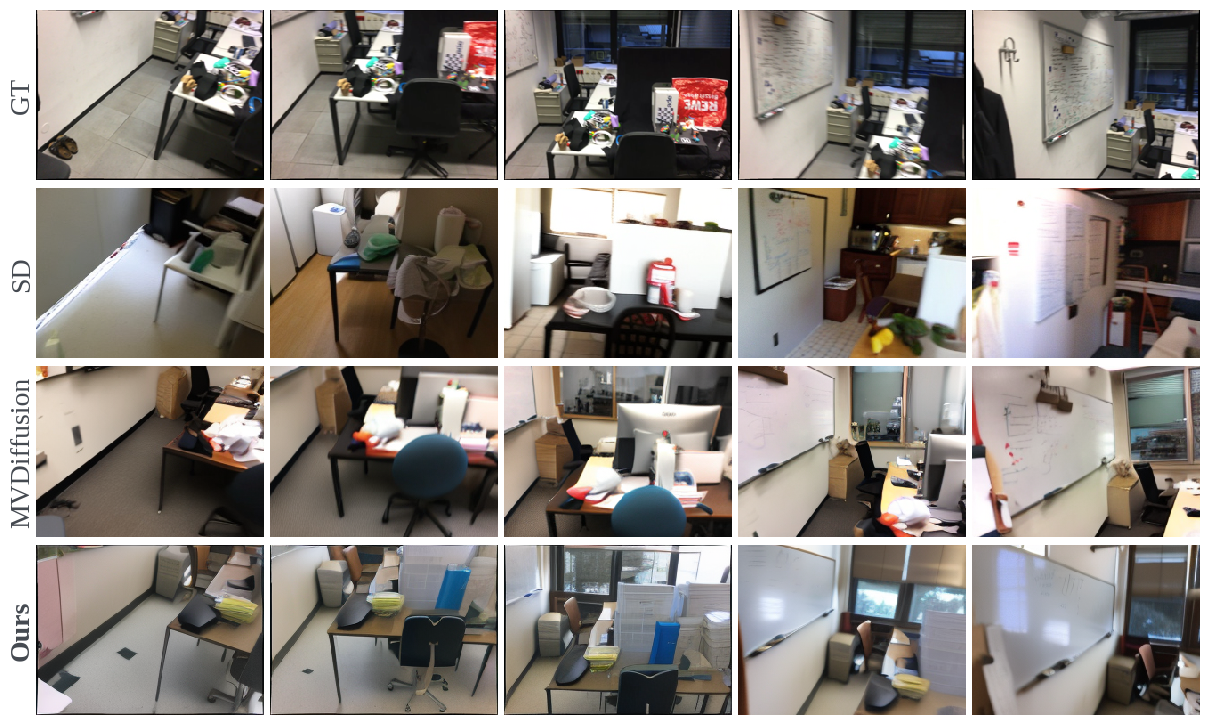}
    \caption{Depth-to-image generation using the prompt ``\textit{a whiteboard on a wall in an office}.''}
    \label{fig:depth_qual2}
\end{figure}

\begin{figure}[h]
    \centering
    \includegraphics[width=0.98\linewidth]{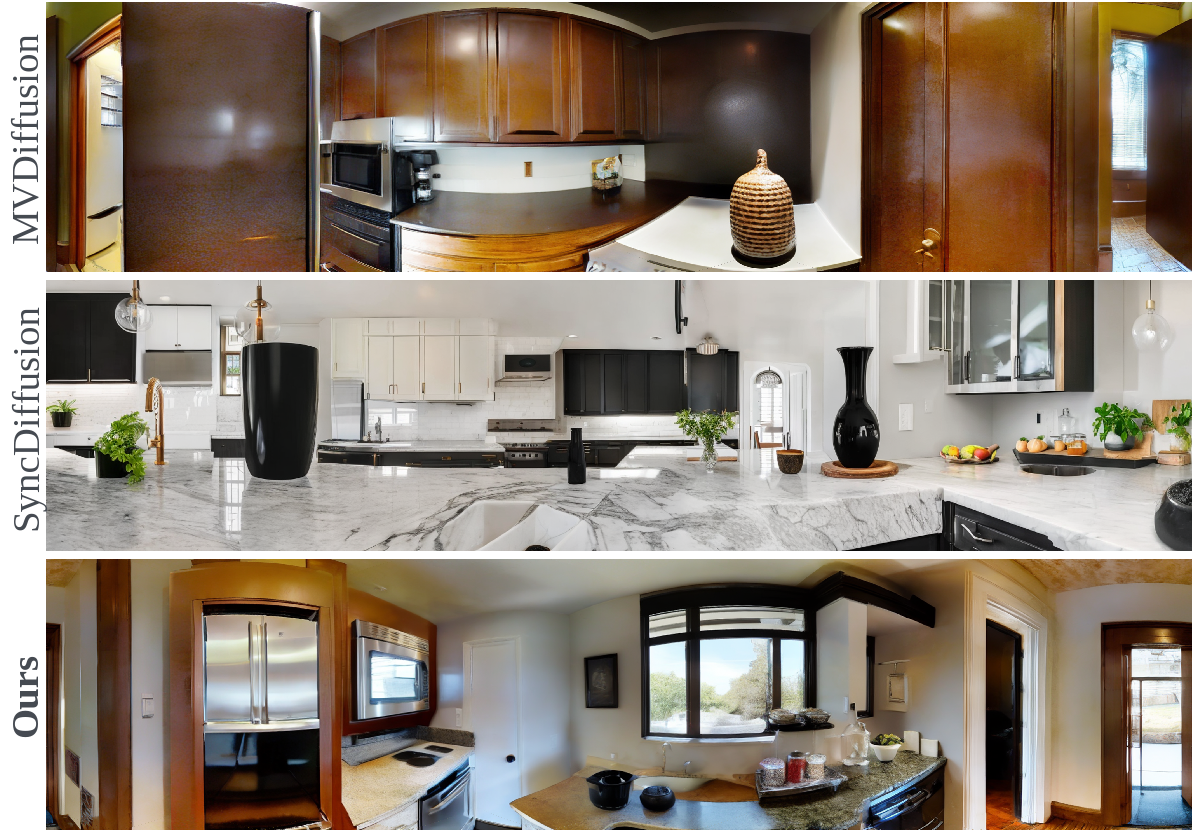}
    \caption{Panoramic image generation using the prompt ``\textit{a kitchen with a large black vase on the counter and a marble counter top next to a sink}.''}
    \label{fig:pano_qual1}
\end{figure}

\begin{figure}[h]
    \centering
    \includegraphics[width=0.98\linewidth]{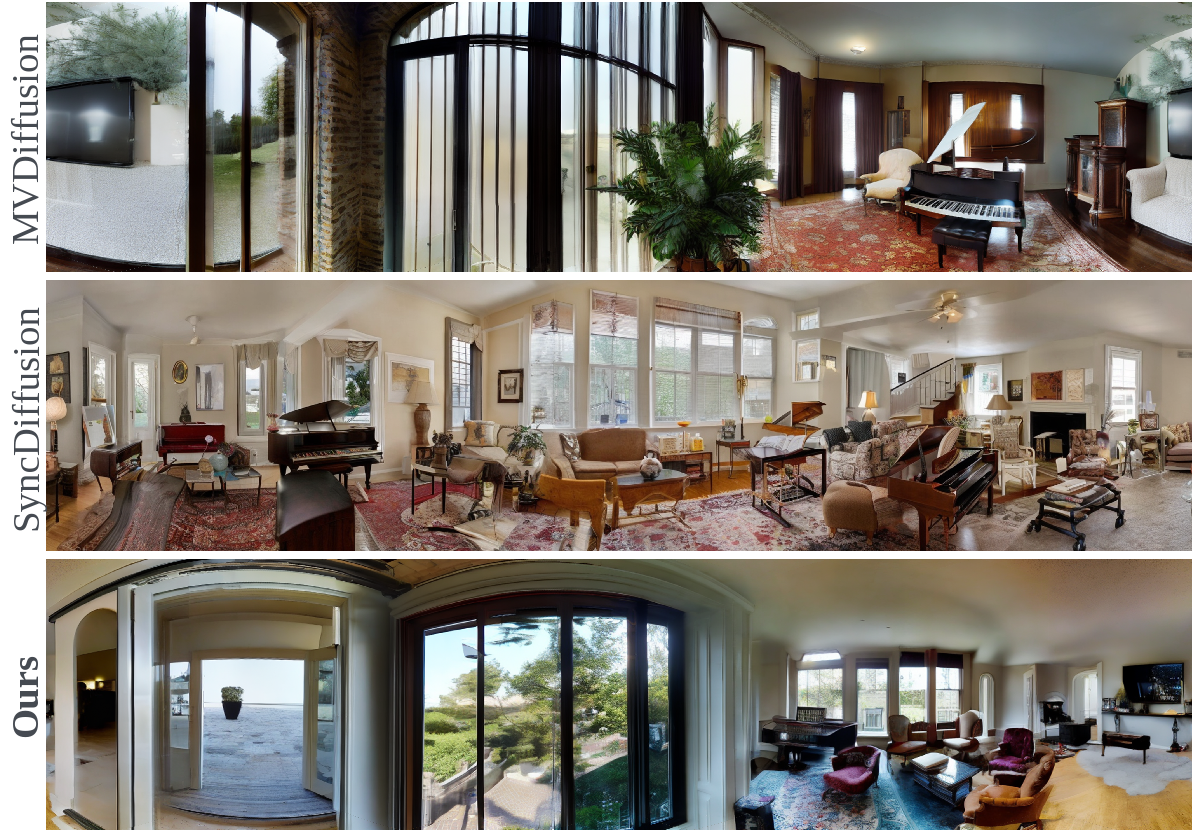}
    \caption{Panoramic image generation using the prompt ``\textit{a living room filled with furniture and a piano}.''}
    \label{fig:pano_qual2}
\end{figure}

\begin{figure}[h]
    \centering
    \includegraphics[width=0.98\linewidth]{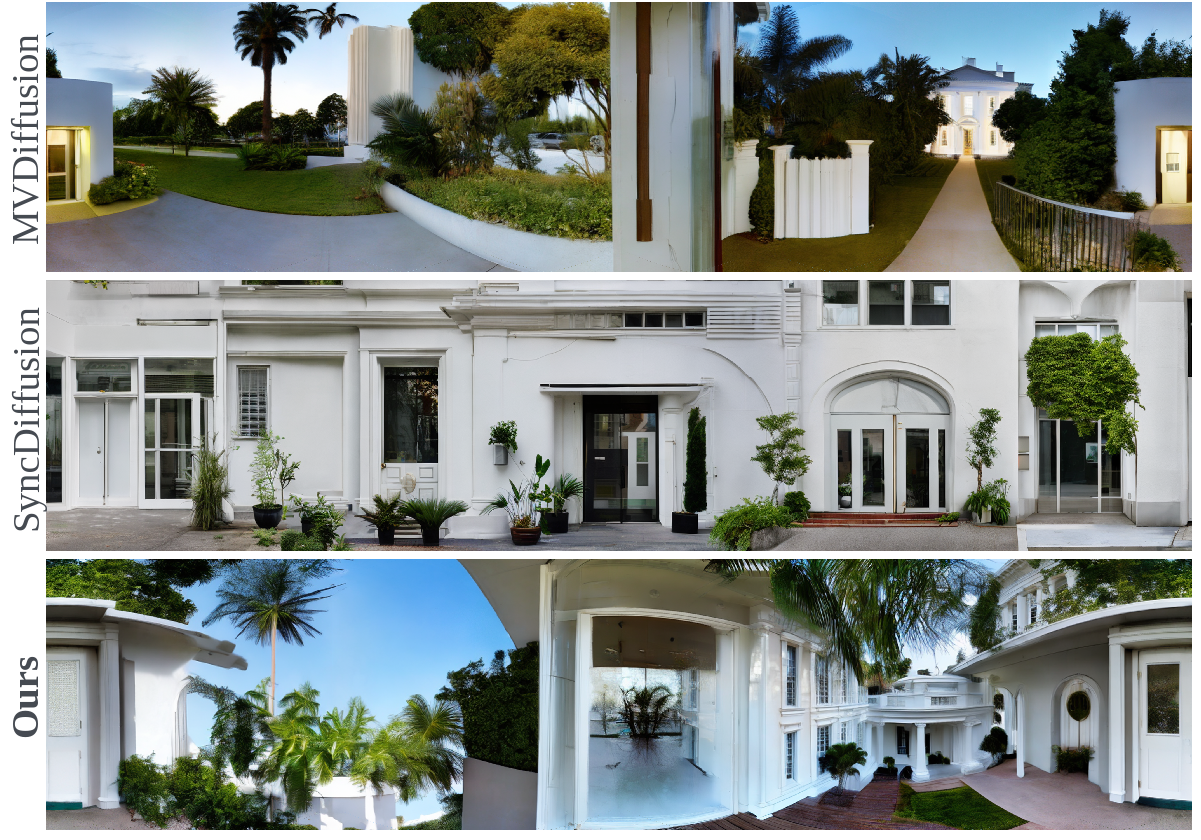}
    \caption{Panoramic image generation using the prompt ``\textit{a white building with a door and some plants in front of a white house with a large glass door}.''}
    \label{fig:pano_qual3}
\end{figure}

\clearpage

%%%%%%%%% REFERENCES
{\small
\bibliographystyle{ieee_fullname}
\bibliography{main}
}

\end{document}